\documentclass[twoside]{article}
\usepackage[accepted]{aistats2eM}
\usepackage{graphicx}
\usepackage{amsmath,amssymb}
\usepackage{multirow}
\usepackage{url}
\usepackage{booktabs}
\usepackage[round]{natbib}
\usepackage{mymath}
\graphicspath{{./}}

\begin{document}

\twocolumn[

\aistatstitle{Beyond Sentiment: The Manifold of Human Emotions}

\aistatsauthor{Seungyeon Kim, Fuxin Li, Guy Lebanon, and Irfan Essa}

\aistatsaddress{ College of Computing \\ Georgia Institute of Technology \\
                 \texttt{\{seungyeon.kim@, fli@cc., lebanon@cc., irfan@cc.\}gatech.edu}} ]

\begin{abstract}
  Sentiment analysis predicts the presence of positive or negative emotions in a text document. In this paper we consider higher dimensional extensions of the sentiment concept, which represent a richer set of human emotions. Our approach goes beyond previous work in that our model contains a continuous manifold rather than a finite set of human emotions. We investigate the resulting model, compare it to psychological observations, and explore its predictive capabilities. Besides obtaining significant improvements over a baseline without manifold, we are also able to visualize different notions of positive sentiment in different domains.
\end{abstract}

\section{Introduction}

Sentiment analysis predicts the presence of a positive or negative emotion $y$ in a text document $x$. Despite its successes in industry, sentiment analysis is limited as it flattens the structure of human emotions into a single dimension. ``Negative'' emotions such as  \texttt{depressed}, \texttt{sad}, and \texttt{worried} are mapped to the negative part of the real line. ``Positive'' emotions such as \texttt{happy}, \texttt{excited}, and \texttt{hopeful} are mapped to the positive part of the real line. Other emotions like \texttt{curious}, \texttt{thoughtful}, and \texttt{tired} are mapped to scalars near 0 or are otherwise ignored. The resulting one dimensional line loses much of the complex structure of human emotions. Note that \emph{emotion}, \emph{affect}, and \emph{mood} have distinguishable meanings in psychology, but we use them here interchangeably.

An alternative that has attracted a few researchers in recent years is to construct a finite collection of emotions and fit a predictive model for each emotion $\{p(y_i|x), i=1,\ldots,C\}$. A multi-label variation that allows a document to reflect more than a single emotion uses a single model $p(y|x)$ where $y\in\{0,1\}^C$ is a binary vector corresponding to the presence or absence of emotions. In contrast to sentiment analysis, this approach models the higher order structure of human emotions.

There are several significant difficulties with the above approach. First, it is hard to capture a complex statistical relationship between a large number of binary variables (representing emotions) and a high dimensional vector (representing the document). It is also hard to imagine a reliable procedure for compiling a finite list of all possible human emotions. Finally, it is not clear how to use documents expressing a certain emotion, for example \texttt{tired}, in fitting a model for predicting a similar one, for example \texttt{sleepy}. Using labeled documents only in fitting models predicting their denoted labels ignores the relationship among emotions, and is problematic for emotions with only a few annotated.

We propose an alternative approach that models a stochastic relationship between the document $X$, an emotion label $Y$ (such as \texttt{sleepy} or \texttt{happy}), and a position on the mood manifold $Z$. We assume that all the emotional aspects in the documents are captured by the manifold, implying that the emotion label $Y$ can be inferred directly from the projection $Z$ of the document on the manifold, without needing to consult the document again.

The key assumption in constructing the manifold Z is that the spatial relationship between $X|Y=j, j=1,\ldots, C $ is similar to the spatial relationship between $Z|Y=j, j=1,\ldots,C$ (see assumption 4 in the next section).

\section{Related Work}\label{sec:related_work}

Studying emotions or affects and their relations is one of the major goals of the psychology community. There are two main approaches: categorical or dimensional. Our focus is on dimensional analysis, as described in \citep{Russell1979,Russell1980,Shaver1987,Watson1985,Watson1988,Tellegen1999,Larsen1992,Yik2011}.

Our work deviates from research in psychology in that we construct our model based on a large collection of annotated documents rather than an experiment with a small number of human subjects. In addition, our model has much higher dimensionality compared to traditional 2-3 dimensions used in psychology.

Sentiment analysis is a significant research direction within the natural language processing community. \citet{Pang2008} is a recent survey of research in this area. Some recent methods are \citep{Nakagawa2010,Socher2011}.

\citet{Alm2008} summarizes affect analysis in text and speech, while \citet{Holzman2003} uses linguistic features to detect emotions in internet chatting. The work described in \citep{Rubin2004,Strapparava2008} classified data using a categorical model suggested by psychological literature. \citet{Mishne2005} and \citet{Genereux2006} examine a similar analysis task using blog posts with standard machine learning techniques, while \citet{Keshtkar2009} exploit a mood hierarchy to improve classification results. The work described in \citep{Strapparava2004,Quan2009,Mohammad2011} address the task of constructing a useful corpus for emotion analysis.

Previous work handles the mood prediction problem as multiclass classification with discrete labels. Our work stands out in that it assumes a continuous mood manifold and thus develops an inherently different learning paradigm. Our logistic regression baseline is generally considered equivalent or better than the ones in related work using SVM \citep{Mishne2005,Genereux2006}, Naive Bayes \citep{Strapparava2008}. \citet{Keshtkar2009} exploited a user-supplied emotional hierarchy which is an additional assumption that we do not have.

\section{The Statistical Model}

We make the following four modeling assumptions concerning the document $X$, the discrete emotion label $Y\in\{1,2,\ldots,C\}$, and the position on the continuous mood manifold $Z\in\mathbb{R}^l$.
\vspace{-1em}
\begin{enumerate}
\item We have the graphical structure: $X \to Z \to Y$, 
implying that the emotion label $Y\in \{1 ,\ldots,C\}$ is independent of the document $X$ given $Z$.
\vspace{-0.5em}
\item The distribution of $Z \in \mathbb{R}^l$ given a specific emotion label $Y=y$ is Gaussian\\
 \parbox[t]{\linewidth}{\vspace{-1.5em}
 \begin{align}
   \label{eq:p_z_gvn_y} \{Z|Y=y\} \sim \mathcal{N}(\mu_y,\Sigma_y).
 \end{align}
}
\vspace{-1em}
\item The distribution of $Z$ given the document $X$ (typically in a bag of words or $n$-gram representation) is a linear regression model\\
  \parbox[t]{\linewidth}{\vspace{-1.5em}
  \begin{align*}
    \{Z|X=x\} \sim \mathcal{N}(\theta^{\top}x,\Sigma_x).
  \end{align*}
 }
\vspace{-1.5em}
\item The distances between the vectors in\\
  \parbox[t]{\linewidth}{\vspace{-1.7em}
  \begin{align*}
    \left\{\E(Z|Y=y): y\in C\right\}
  \end{align*}
 } are similar to the corresponding distances in\\
  \parbox[t]{\linewidth}{\vspace{-1.7em}
  \begin{align*}
    \left\{\E(X|Y=y): y\in C\right\}
  \end{align*}
 }
\vspace{-2em}
\end{enumerate}
We make the following observations.\vspace{-1em}
\begin{itemize}
  \item The first assumption implies that the emotion label $Y$ is simply a discretization of the continuous $Z$. It is consistent with well known research in psychology (see Section 2) and with random projection theory, which state that it is often possible to approximate high dimensional data by projecting it on a low dimensional continuous space. \vspace{-0.5em}
  \item While $X$, $Y$ are high dimensional and discrete, $Z$ is low dimensional and continuous. This, together with the conditional independence in assumption (1) above, implies a higher degree of accuracy than modeling directly $X\to Y$. Intuitively, the number of parameters is on the order of $\text{dim}(X)+\text{dim}(Y)$ as opposed to $\text{dim}(X)\text{dim}(Y)$. \vspace{-1.5em}
  \item The Gaussian models in assumptions 2 and 3 are simple, and lead to efficient computational procedures. We also found them to work well in our experiments. The model may be easily adapted, however, to more complex models such as mixture of Gaussians or non-linear regression models (for example, we experimented with quadratic regression models). \vspace{-0.5em}
  \item Assumption 4 suggests that we can estimate $\E(Z|Y=y)$ for all $y\in C$ via multidimensional scaling. MDS finds low dimensional coordinates for a set of points that approximates the spatial relationship between the points in the original high dimensional space. \vspace{-0.5em}
  \item The models in assumptions 2 and 3 are statistical and can be estimated from data using maximum likelihood. \vspace{-0.5em}
  \item The four assumptions above are essential in the sense that if any one of them is removed, we will not be able to consistently estimate the true model. \vspace{-0.5em}
\end{itemize}

\begin{figure}
  \centering
  \includegraphics[width=.75\linewidth,trim=1.5em 0 1.5em 2.5em,clip]{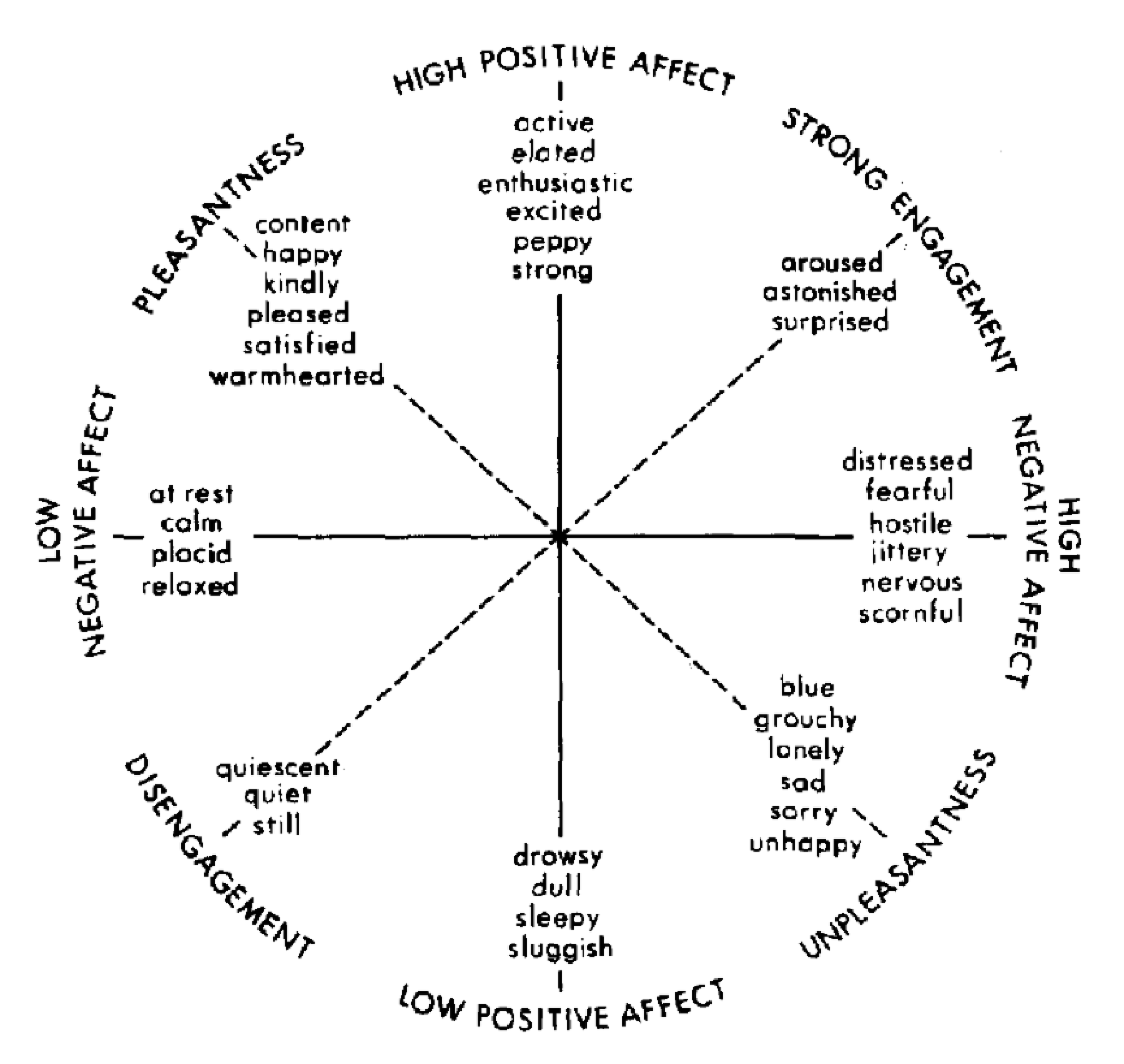}
  \vspace{-0.5em}
  \caption{The two-dimensional structure of emotions from \citep{Watson1985}.  We can interpret top-left to bottom-right axis as expressing sentiment polarity and the top-right to bottom-left axis as expressing engagement.}\label{fig:watson}
  \vspace{-0.5em}
\end{figure}

\subsection{Fitting Parameters and Using the Model} \label{sec:fitting}

Motivated by the fourth modeling assumption, we determine the parameters $\mu_y=\E(Z|Y=y), y\in C$ by running multidimensional scaling (MDS) or Kernel PCA on the empirical versions of $\{\E(X|Y=y): y\in C\}$, which are the class averages $\frac{1}{n_k} \sum_{y^{(i)} = k} x^{(i)}$ ($n_k$ is the number of documents belonging to category $k$).

We estimate the parameter $\theta$, defining the regression $X\to Z$, by maximizing the likelihood
\begin{align} \label{eq:mle1}
  \hat{\theta} &= \argmax_{\theta} \sum_i \log p(y^{(i)}| x^{(i)})  \\
          &= \argmax_{\theta} \sum_i \log \int_Z p(y^{(i)}| z) p_{\theta}(z|x^{(i)}) dz  \nonumber \\
          &= \argmax_{\theta} \sum_i \log \int_Z p(z|y^{(i)}) \frac{p(y^{(i)}) p_{\theta}(z|x^{(i)})}{\sum_y p(z|y) p(y)} dz.\nonumber
\end{align}

The covariance matrices $\Sigma_y$ of the Gaussians $Z|Y=y$, $y=1,\ldots,C$ may be estimated by computing the empirical variance of $Z$ values simulated from $p_{\hat\theta}(Z|X^{(i)})$
for all documents $X^{(i)}$ possessing the right labels $Y^{(i)}=y$. A more computationally efficient alternative is computing the empirical variance of the most likely $\hat{Z}^{(i)}$ values corresponding to documents possessing the appropriate label $Y^{(i)}=y$:
\begin{align}
  \hat{Z}^{(i)} = \argmax_z p_{\hat\theta}(Z=z|X^{(i)})
    =\hat\theta^{\top} X^{(i)}.\label{eq:zhat}
\end{align}

Given a new test document $x$, we can predict the most likely emotion with
\begin{align}\nonumber
  \hat{y} &= \argmax_y \int p(y,z|x)dz \notag \\
          &= \argmax_y \int p(y|z)p_{\hat{\theta}} (z|x) dz. \label{eq:predict}
\end{align}

But in many cases, the distribution $p(Z|X)$ provides more insightful information than the single most likely emotion $Y$.

\subsection{Approximating High Dimensional Integrals}

Some of the equations in the previous section require integrating over $Z\in\R^l$, a computationally difficult task when $l$ is not very low. There are, however, several ways to approximate these integrals in a computationally efficient way.

The most well-known approximation is probably Markov chain Monte Carlo (MCMC). Another alternative is the Laplace approximation. A third alternative is based on approximating the Gaussian pdf with Dirac's delta function, also known as an impulse function,  resulting in the approximation
\begin{align} \nonumber
\int N(z\,;\mu,\Sigma) g(z)\, dz &\approx
c(\Sigma)\int \delta(z-\mu)g(z)\,dz \\ &=
c(\Sigma) g(\mu). \label{eq:approx}
\end{align}
A similar approximation can also be derived using Laplace's method. Obviously, the approximation quality  increases as the variance decreases.

Applying \eqref{eq:approx} to \eqref{eq:mle1} we get \vspace{-1em}
\begin{align} \nonumber
\hat\theta   &\approx \argmax_{\theta} \sum_i \log \frac{p(y^{(i)}) p_{\theta}({z^{(i)}}^* |x^{(i)})}{\sum_y p({z^{(i)}}^*|y) p(y)} \label{eq:dirac_delta_approx} \\
&= \argmax_{\theta} \sum_i \log p_{\theta}({z^{(i)}}^* |x^{(i)})\vspace{-0.5em}
\end{align}
where ${z^{(i)}}^* = \argmax_z p(z|y^{(i)}) = E(Z|y^{(i)})$,
which is equivalent to a least squares regression.

Applying \eqref{eq:approx} to \eqref{eq:predict} yields a classification rule
\begin{align}
\hat{y}  &\approx \argmax_y p\left(y\Big|Z=\argmax_z p_{\hat\theta}(z|x)\right). \label{eq:map_approx}
\end{align}

\subsection{Implementation} \label{sec:impl-detail}

In estimating the covariance matrices of a Gaussian $P(Z|Y=y)$, it is sometimes assumed that each class has the same covariance matrix, leading to linear discriminant analysis (LDA) as the optimal Bayes classifier. The alternative assumption that the covariance matrices for each class is different leads to quadratic discriminant analysis (QDA) as the optimal Bayes classifier. 

We consider both assumptions and three different models for the covariance matrices: full covariance, diagonal covariance, and linear combination of full covariance and spherical covariance (standard regularization technique):\vspace{-0.5em}
\begin{align*}
\hat{\Sigma}' =& (1-\lambda)  \hat{\Sigma} + \lambda  \left( \sum_{i=1}^C \hat\Sigma_{ii} \right)I \quad &\text{(LDA)} \\ 
  \hat{\Sigma}_{y}' =& (1-\lambda) \hat{\Sigma}_y + \lambda \left( \sum_{i=1}^C [\hat{\Sigma}_y]_{ii} \right) I \quad &\text{(QDA)}.
\end{align*} 
In either case we used a $C$ dimensional ambient space ($C$ equals the number of emotions) and the approximation \eqref{eq:map_approx}.

Due to the high dimensionality of $X$, it may be useful to estimate $\hat\theta$ using ridge regression, rather than least squares regression. In this case, we update the estimate $E(Z|Y=y)$ in third stage, based on the ridge estimate $\hat\theta$.

One interpretation of our model $X\to Z\to Y$ is that $Z$ forms a sufficient statistic of $X$ for $Y$. We can thus consider adapting a wide variety of predictive models (for example, logistic regression or SVM) on $Z\mapsto Y$. These discriminative classifiers are trained on $\{(\hat{Z}^{(i)},Y^{(i)}), i=1,\ldots,n\}$.

\section{Experiments}

\subsection{Datasets}

We used crawled Livejournal\footnote{\url{http://www.livejournal.com}} data as the main dataset. Livejournal is a popular blog service that offers emotion annotation capabilities to the authors. About 20\% of the blog posts feature these optional annotations in the form of emoticons. The annotations may be chosen from a pre-defined list of possible emotions, or a novel emotion specified by the author. We crawled 15,910,060 documents and selected 1,346,937 documents featuring the most popular 32 emotion labels (in respect to the number of documents annotated in). It is a significantly larger dataset compare to similar works: 1,000 \citep{Strapparava2008}, 346,723 \citep{Genereux2006} and 345,014 \citep{Mishne2005} documents.

We used Indri from the Lemur project\footnote{\url{http://www.lemurproject.org/}} to extract term frequency features while tokenizing and stemming (using the Krovetz stemmer) words. As is common in sentiment studies \citep{Das2007,Na2004,Kennedy2006} we added new features representing negated words. For example, the phrase ``not good'' is represented as a token ``not-good'' rather than as two separate words. This resulted in 43,910 features.

We used $L_1$-normalization, dividing term frequency matrix by the number of total word appearances in each document, and followed with a square root transformation, turning the Euclidean distance to the Hellinger distance. This multinomial geometry outperforms the Euclidean geometry in a variety of text processing tasks, as described in \citep{Lafferty2005a,Lebanon2005b}.

Building a model solely based on the engineered term frequency features ignores the structure of a sentence or paragraphs. Using richer sets of feature may improve our model further; however, our contribution is presenting the manifold of emotions. We will use richer feature, especially handling sentence structures, in later research.

The document length histogram is close to an exponential distribution, with mean 113.51 words and standard deviation 146.65 words. There are plenty of short documents (520,436) having less than $50$ words, but there are also some long documents (39,570) having more than 500 words. The average word length is 8.33 characters.

Two other datasets that we use in our experiments are the movie review data \citep{Pang2005} and the restaurant review data\footnote{\url{http://www.cs.cmu.edu/~mehrbod/RR/}} \citep{Ganu2009} (using the same preprocessing described above).

\begin{figure}
  \includegraphics[width=\linewidth,trim=0.3em 0 0.6em 0,clip]{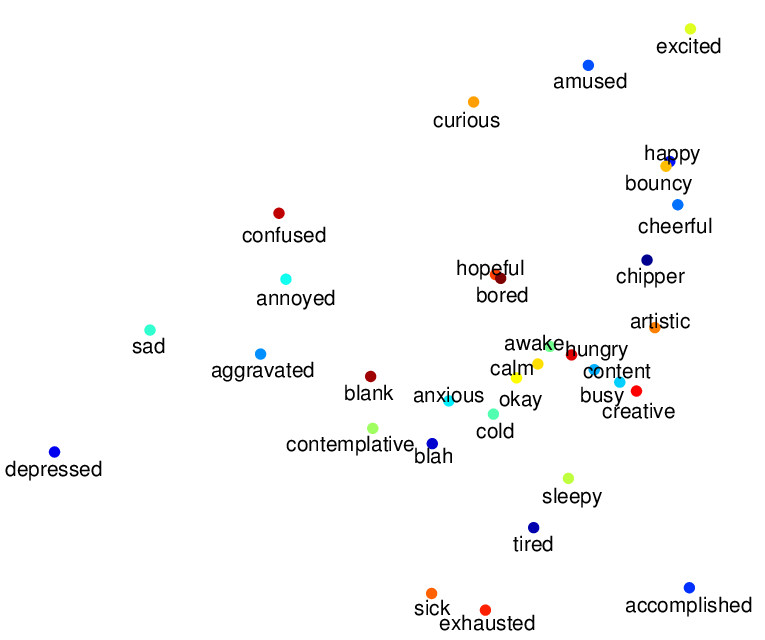}
  \vspace{-2em}
  \caption{Mood centroids $\E(Z|Y=y)$ on the two most prominent dimensions in emotion space fitted from blog posts. The horizontal dimension corresponds to sentiments polarity and the vertical dimension corresponds to mental engagement level (compare with Figure.~\ref{fig:watson}).}\label{fig:2d_embedding}
  \vspace{-1em}
\end{figure}

\begin{figure}
  \begin{center}
  \includegraphics[height=.7\linewidth,trim=1em 0em 1em 0,clip,angle=90]{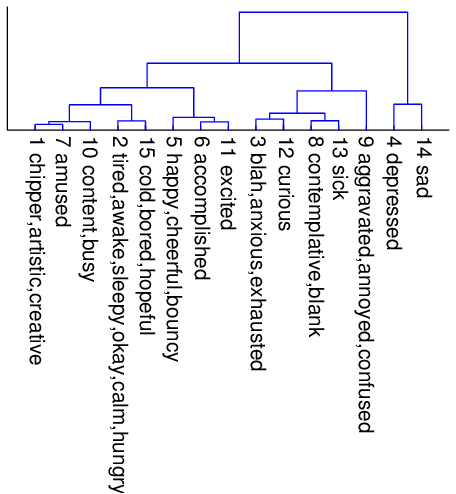}
  \end{center}
  \vspace{-1em}
  \caption{Dendrogram of moods using complete linkage function on Bhattacharyya distances between moods. The leaves are cut in 15 clusters to reduce clutters.}\label{fig:dendrogram}
  \vspace{-1em}
\end{figure}

\subsection{Comparison with Psychological Models}

In this section, we compare our model to Watson and Tellegen's well known psychological model (Figure~\ref{fig:watson}). Figure~\ref{fig:2d_embedding} shows the locations of mood centroids $\E(Z|Y=y)$ on the first two dimensions of the mood manifold.  We make the following observations.
\vspace{-0.5em}
\begin{enumerate}
  \item The horizontal axis expresses a sentiment polarity-like emotion. The left part features emotions such as \texttt{sad} and \texttt{depressed}, while the right part features emotions such as \texttt{accomplished}, \texttt{happy} and \texttt{excited}. This is in agreement with Watson and Tellegen's observations (see Figure~\ref{fig:watson}) that identify sentiment polarity as the most prominent factor among human emotions.\vspace{-0.5em}    \item The vertical axis expresses the level of mental engagement or energy level. The top part features emotions such as \texttt{curious} or \texttt{excited}, while the bottom part features emotions such as \texttt{exhausted} or \texttt{tired}. This agrees partially with the engagement dimension in the psychological model. However, the precise definition of engagement seems to be different. For example, in our model (Figure~\ref{fig:2d_embedding}), high engagement imply active conscious mental states, such as \texttt{curious}, rather than passive emotions such as \texttt{astonished} and \texttt{surprised} (Figure~\ref{fig:watson}).\vspace{-0.5em}
  \item The neutral moods \texttt{blank}, stay in the middle of the picture.\vspace{-0.5em}
  \end{enumerate}

The mood centroid figure is largely intuitive, but the positions of a few centroids is somewhat unintuitive; for example \texttt{annoyed} has similar vertical location (energy level) as \texttt{bored}. We note, however, the manifold is higher dimensional and the dimensions beyond the first two provide additional positioning information.

It is interesting to consider the list of words that are most highly scored for each axis in our mood manifold. The words with highest weight associated with the horizontal axis (sentiment polarity) are: \texttt{depress, sad, hate, cry, fuck, sigh, died} on the left (negative) side and \texttt{excite, awesome, yay, happy, lol, xd, fun} on the right (positive) side. On the vertical axis (energy): \texttt{tire, download, exhauste, sleep, sick, finishe, bed} on the bottom side (low energy) and \texttt{excite, amuse, laugh, not-wait, hilarious, curious, funny} on the top side (high energy).

We conclude that there is in large part an agreement between the first two dimensions in our model and the standard psychological model. This agreement between our mood manifold and the psychological findings is remarkable in light of the fact that the two models used completely different experimental methodology (blog data vs. surveys).

\begin{figure}
  \centering
  \includegraphics[height=.8\linewidth,trim=3.9em 0.5em 0.5em 1.8em,clip]{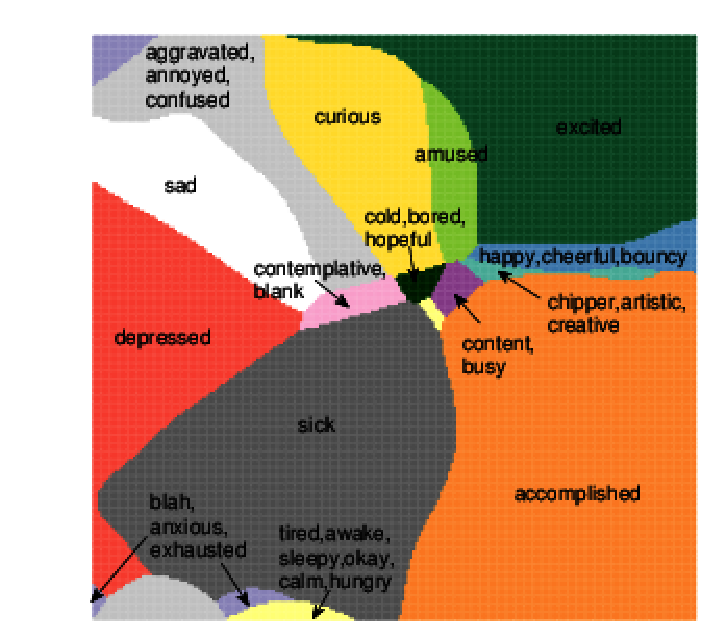}
  \vspace{-1em}
  \caption{Tessellation of the space spanned by the first two dimensions of mood manifold with 15 ``super-emotion'' clusters ($\argmax_y p(Z|Y=t)$). }\label{fig:cluster_partition}
  \vspace{-1em}
\end{figure}

\begin{table*}  
  \caption{Macro F1 score and accuracy over the test set in multiclass emotion classification over top 32 moods(left) and over 7 clusters from Figure~\ref{fig:dendrogram}(right). Bold text represent statistically significant  ($t$-test) improvement by using the mood manifold over the corresponding classification method in the original feature space.} \label{tbl:classification_mood}
  \footnotesize
  \begin{tabular*}{.48\linewidth}{@{\extracolsep{\fill}} l l r r r r}
    \toprule
    &  & \multicolumn{2}{c}{Original Space} &  \multicolumn{2}{c}{Mood Manifold}\\
    &  &  F1 & Acc. & F1 & Acc. \\
    \midrule
    LDA & full   &    n/a &    n/a & \bf 0.1247 & \bf 0.1635 \\
        & diag.  & \bf 0.1229 & 0.1441 & 0.1160 & \bf 0.1600 \\
        & spher. & 0.0838 & 0.1075 & \bf 0.0896 & \bf 0.1303 \\
    QDA & full   &    n/a &    n/a & \bf 0.1206 & \bf 0.1478 \\
        & diag.  & 0.0878 & 0.0931 & \bf 0.1118 & \bf 0.1463 \\
        & spher. & 0.0777 & 0.0989 & \bf 0.0873 & \bf 0.1253 \\
    Log.Reg. &   & 0.1231 & 0.1360 & \bf 0.1477 & \bf 0.1667 \\
    \bottomrule
  \end{tabular*}
  \quad
  \begin{tabular*}{.48\linewidth}{@{\extracolsep{\fill}} l l r r r r}
    \toprule
    &  & \multicolumn{2}{c}{Original Space} &  \multicolumn{2}{c}{Mood Manifold}\\
    &  &  F1 & Acc. & F1 & Acc. \\
    \midrule
    LDA & full   &    n/a &    n/a & \bf 0.2800 & \bf 0.3591 \\
        & diag.  & 0.2661 & 0.2890 & \bf 0.2806 & \bf 0.3504 \\
        & spher. & 0.2056 & 0.2252 & \bf 0.2344 & \bf 0.2876 \\
    QDA & full   &    n/a &    n/a & \bf 0.2506 & \bf 0.3025 \\
        & diag.  & 0.1869 & 0.1918 & \bf 0.2496 & \bf 0.3088 \\
        & spher. & 0.1892 & 0.2009 & \bf 0.2332 & \bf 0.2870 \\
    Log.Reg. &   & \bf 0.2835 & 0.3459 & 0.2806 & \bf 0.3620 \\
    \bottomrule
  \end{tabular*}
  \vspace{-1em}
\end{table*}

\subsection{Exploring the Emotion Space}

Since emotion labels correspond to distributions $P(Z|Y)$, we can cluster these distribution in order to analyze the relationship between the different emotion labels. In the first analysis, we perform hierarchical clustering on the emotions in order to create emotional concepts of varying  granularity. This is especially helpful when the original emotions are too fine, (consider for example the two distinct but very similar emotions \texttt{annoyed} and \texttt{aggravated}). In the second analysis we visualize the 2D tessellation corresponding to most likely emotions in mood space. This reveals additional information, beyond the centroid locations in Figure \ref{fig:2d_embedding}. 

We use the  
Bhattacharyya dissimilarity, \vspace{-0.7em}
\begin{align*}
D_B(f,g)  &= -\log \int\sqrt{f(z) g(z)} dz.
\vspace{-0.5em}
\end{align*}
to measure dissimilarity between emotions, which corresponds to the log Hellinger distance between the underlying distributions. In the case of two multivariate Gaussians, it has the following closed form: \vspace{-0.7em}
\begin{multline*}
  D_B(N(\mu_1,\Sigma_1),N(\mu_2,\Sigma_2)) \\ = \frac{1}{8}(\mu_1 - \mu_2)^T \left(\frac{\Sigma_1 + \Sigma_2}{2}\right)^{-1} (\mu_1 - \mu_2) \\
      + \frac{1}{2} \log \left( \frac{\det((\Sigma_1 + \Sigma_2)/2)}{\sqrt{\det \Sigma_1 \det \Sigma_2}} \right).
\end{multline*}
Following common practice, we add a small value to the diagonal of the covariance matrices to ensure invertibility.

Figure~\ref{fig:dendrogram} shows the mood dendrogram obtained by hierarchical clustering of the top 32 emotions using the Bhattacharyya dissimilarity (complete linkage clustering). The bottom part of dendrogram was omitted due to lack of space. The clustering agrees with our intuition in many cases. For example,
\vspace{-0.5em}
\begin{enumerate}
  \item \texttt{aggravated,annoyed} and \texttt{confused} are in the same tight cluster.\vspace{-0.5em}
  \item \texttt{sad} and \texttt{depressed} are very close cluster.\vspace{-0.5em}
  \item \texttt{happy}, \texttt{cheerful}, and \texttt{bouncy} are in the same tight cluster, which is close to \texttt{accomplished} and \texttt{excited}.\vspace{-0.5em}
  \item \texttt{tired, awake, sleepy, okay, calm} and \texttt{hungry} are in the same tight cluster.\vspace{-0.5em}
\end{enumerate}

The hierarchical clustering is useful in aggregating similar emotions. If the situation requires paying attention to one or two ``types'' of emotions, we can use a particular mood cluster to reflect the desired feature. For example, when analyzing product reviews we may want to partition the emotions into two clusters: positive and negative. When analyzing the effect of a new advertisement campaign we may be interested in a clustering based on positive engagement: excited / energetic vs. bored. Other situations may call for other clusters of emotions.

Figure~\ref{fig:cluster_partition} shows the tessellation corresponding to \[f(z)=\argmax_{y=1,\ldots,C} p(Z|Y=y).\]
For space and clarity purposes, we use 15 emotion clusters instead of the entire set of 32 emotions. The tessellation shows the regions being classified to each emotion cluster based only on the 2D space. We observe that:
\vspace{-0.5em}
\begin{enumerate}
  \item As in Figure~\ref{fig:2d_embedding} the horizontal axis corresponds to negative(left) - positive(right) emotion and the vertical axis corresponds to energy level(or engagement): (top) \texttt{excited} and \texttt{curious} vs. (bottom) \texttt{tired} and \texttt{exhausted}. \vspace{-0.5em}
  \item The \texttt{depressed} region is spread significantly on the left-bottom side, and is neighboring the \texttt{sick} region and the \texttt{sad} region. \vspace{-0.5em}
  \item  The region corresponding to the \texttt{happy, cheerful, bouncy} emotions neighbors the  \texttt{accomplished} region and the \texttt{excited} region. \vspace{-0.5em}
\end{enumerate}

A similar tessellation of a higher dimensional $Z$ space provides additional information. However, visualizing such higher dimensional spaces is substantially harder in paper format.

\begin{table*}
  \caption{F1 and accuracy over test-set in  sentiment polarity task (left): \{cheerful, happy, amused\} vs \{sad, annoyed, depressed, confused\}, and detecting energy level (right) \{sick, exhausted, tired\} vs. \{curious, amused\}. Bold text represent statistically significant  ($t$-test) improvement by using the mood manifold over the corresponding classification method in the original feature space.} \label{tbl:classification_binary}
  \footnotesize
  \begin{tabular*}{.48\linewidth}{@{\extracolsep{\fill}} l l r r r r}
    \toprule
    &  & \multicolumn{2}{c}{Original Space} &  \multicolumn{2}{c}{Mood Manifold}\\
    &  &  F1 & Acc. & F1 & Acc. \\
    \midrule
    LDA & full   &    n/a &    n/a & \bf 0.7340 & \bf 0.7812 \\
        & diag.  & 0.7183 & 0.7436 & \bf 0.7365 & \bf 0.7663 \\
        & spher. & 0.6358 & 0.6553 & \bf 0.7482 & \bf 0.7699 \\
    QDA & full   &    n/a &    n/a & \bf 0.6500 & \bf 0.7446 \\
        & diag.  & 0.6390 & 0.6398 & \bf 0.6704 & \bf 0.7510 \\
        & spher. & 0.6091 & 0.6143 & \bf 0.7472 & \bf 0.7734 \\
    Log.Reg. &   & 0.7350 & 0.7624 & \bf 0.7509 & \bf 0.7857\\
    \bottomrule
  \end{tabular*}
  \quad
  \begin{tabular*}{.48\linewidth}{@{\extracolsep{\fill}} l l r r r r}
    \toprule
    &  & \multicolumn{2}{c}{Original Space} &  \multicolumn{2}{c}{Mood Manifold}\\
    &  &  F1 & Acc. & F1 & Acc. \\
    \midrule
    LDA & full   &    n/a &    n/a & \bf 0.7084 & \bf 0.7086 \\
        & diag.  & 0.6441 & 0.6449 & \bf 0.6987 & \bf 0.6989 \\
        & spher. & 0.6343 & 0.6343 & \bf 0.6913 & \bf 0.6913 \\
    QDA & full   &    n/a &    n/a & \bf 0.5706 & \bf 0.6100 \\
        & diag.  & 0.6124 & 0.6413 & \bf 0.6268 & \bf 0.6446 \\
        & spher. & 0.6239 & 0.6294 & \bf 0.6754 & \bf 0.6767 \\
    Log.Reg. &   & 0.6694 & 0.6699 & \bf 0.7087 & \bf 0.7089 \\
    \bottomrule
  \end{tabular*}
  \vspace{-1em}
\end{table*}

\subsection{Classifying Emotions}

One of the primary experiment in this paper is emotion classification. In other words, given a document $x$ predict the emotion that is expressed in the text. As mentioned in the introduction, this classification can be done by constructing separate $p(y_i|x)$ models for every emotion (one-vs-all approach). However, the one vs. all approach is not entirely satisfactory as it ignores the relationships between similar and contradictory moods. For example, documents labeled as \texttt{sleepy} can be helpful when we fit a model for predicting \texttt{tired}. The mood manifold provides a natural way to incorporate this information, as documents from similar moods will be mapped to similar points on the manifold.

Besides testing different variants of LDA and QDA, we also compare logistic regression on the original input space and on the mood manifold (see Section~\ref{sec:impl-detail}).

\subsubsection{Experiment Details}

We performed emotion classification experiment (Table~\ref{tbl:classification_mood}, left) on the Livejournal data. We considered the goal of predicting the most popular 32 moods. The class proportion varies in the range 1.72\% to 6.52\%.

Since 32 moods are too finer in practical usage, we designed coarser classification experiment (Table~\ref{tbl:classification_mood}, right) using 7 clusters obtained by hierarchical clustering as in Figure~\ref{fig:dendrogram}. The task is to predict the 7 clusters and cluster proportion varies in the range 4.02\% to 28.63\%.

We also considered two binary classification tasks (Table~\ref{tbl:classification_binary}) obtained by partitioning the set of moods into two clusters (positive vs. negative clusters and high vs. low energy clusters). The class distributions of these binary tasks are 65.03\% vs. 34.97\% (sentiment polarity), and 52.17\% vs. 47.83\% (energy level)

We used half of the data for training and half for testing. To determine statistical significance, we performed $t$-tests on several random trials. Note that emotion prediction is a hard task, as similar emotions are hard to discriminate (consider for example discriminating between \texttt{aggravated} and \texttt{annoyed}). It is thus not surprising that prediction performances are relatively low, especially when discriminating between a large number of moods or clusters. 

The LDA, QDA and $L_2$-regularized logistic regression models are implemented in MATLAB (the latter with LBFGS solver). We also regularized the LDA and QDA models by considering multiple models for the covariance matrices. We determined the regularization parameters by examining the performance of the model (on a validation set) on a grid of possible parameter values. We used the same parameters in all our experiments.

\begin{figure}
  \hspace{1em}
  \includegraphics[width=.75\linewidth,trim=0em 0em 0em 1.5em,clip]{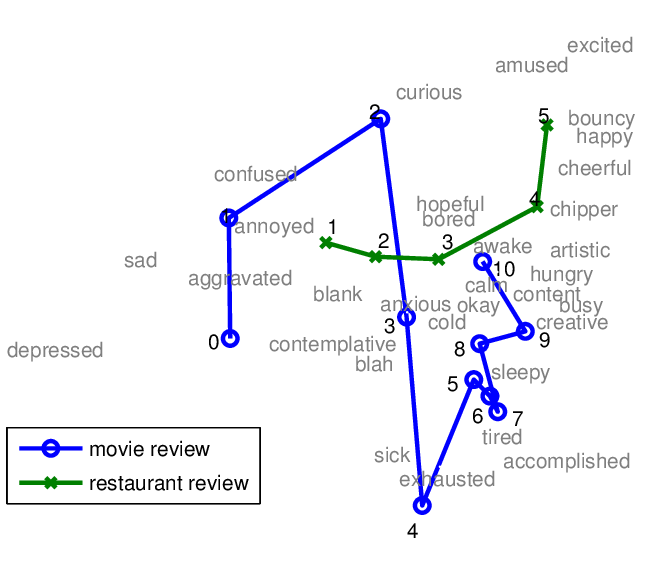}
  \vspace{-2em}
  \caption{Projected centroids of each review score (higher is better) of movie reviews and restaurant reviews on the mood manifold. Both review start from the left side (negative sentiment in mood manifold) and continues to the right side (positive sentiment) with two different unique patterns.}\label{fig:sentiment_notion}
  \vspace{-1em}
\end{figure}

\subsubsection{Classification Results}

Table~\ref{tbl:classification_mood} and \ref{tbl:classification_binary} compare classification results using the original bag of words feature space and the manifold model, using different types of classification methods: LDA, QDA with different covariance matrix models, and logistic regression. Bold faces are improvements over the baseline with statistical significance of $t$-test of random trials.

Most of experimental results show that the mood manifold model results in statistically significant improvements than using original bag of words feature. Improvements are consistent with various choices of classification methods: LDA, QDA, or logistic regression. The phenomenon is also persistent in variety of tasks: 32 mood classification, more practical 7 cluster classification, or binary tasks. Thus, introducing the mood manifold is indeed made the difference.

\begin{figure*}
  \centering
  \includegraphics[width=.24\linewidth,trim=0.2em 0.2em 0.4em 0,clip]{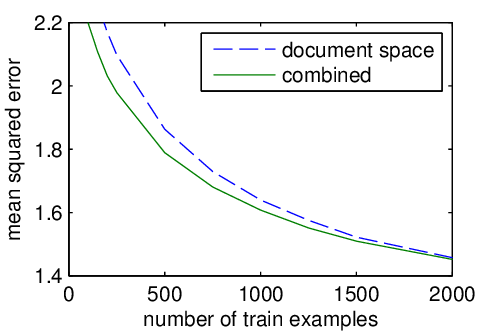}
  \includegraphics[width=.24\linewidth,trim=0.2em 0.2em 0.4em 0,clip]{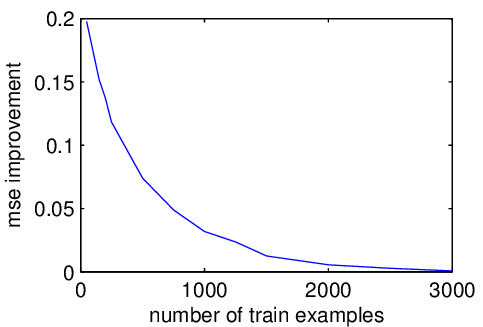}
  \includegraphics[width=.24\linewidth,trim=0.2em 0.2em 0.4em 0,clip]{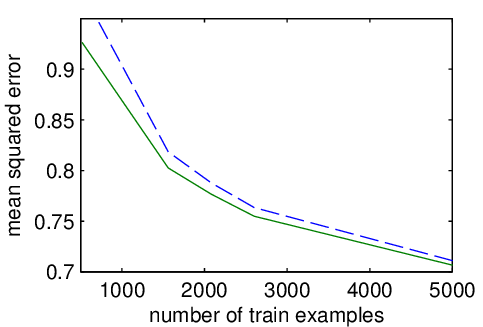}
  \includegraphics[width=.24\linewidth,trim=0.2em 0.2em 0.4em 0,clip]{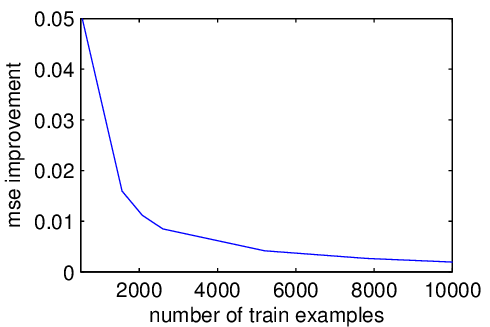}
\vspace{-1em}
\caption{Test set mean squared error and its improvements on movie review (two figures on the left) and restaurant review (two figures on the right) as a function of the sentiment train set size. Prediction using the combined features outperforms the baseline (regression on document space) and the advantage is larger on smaller training set.}
  \label{fig:review_prediction}
\vspace{-1.5em}
\end{figure*}

\section{Application}

\subsection{Improving Sentiment Prediction using Mood Manifold} \label{sec:other_sentiment_notions}

The concept of positive-negative sentiment fits naturally within our framework as it is the first factor in the continuous $Z$ space. Nevertheless, it is unlikely that all sentiment analysis concepts will align perfectly with this dimension. For example, movie reviews and restaurant reviews do not represent identical concepts. In this subsection we visually explore these concepts on the manifold and  show that the mood manifold leads to improved sentiment polarity prediction on these domains.

\subsubsection{Sentiment Notion on the Manifold}
We model a sentiment polarity concept as a smooth one dimensional curve within the continuous $Z$ space. As we traverse the curve, we encounter  documents corresponding to negative sentiments, changing smoothly into emotions corresponding to positive sentiments. We complement the stochastic embedding $p(Z|X)$ with a smooth probabilistic mapping $\pi(R|Z)$ into the sentiment scale. The prediction rule becomes \vspace{-0.5em}
\[ \hat r = \argmax_r \int p(Z=z|X) \pi(R=r|Z=z)\,dz\]
and its approximated version is \vspace{-0.5em}
\[ \hat r = \argmax_r \pi\left(R=r\Big|Z=\argmax_z P(Z=z|X)\right)\]

Figure~\ref{fig:sentiment_notion} shows the smooth curves corresponding to $\E\left[\pi(R=r|Z)\right]$ for movie reviews and restaurant reviews. Both curves progress from the left (negative sentiment) to the right (positive sentiment). But the two curves show a clear distinction: the movie review sentiment concept is in the bottom part of the figure, while the restaurant review sentiment concept is in the top part of the figure. We conclude that positive restaurant reviews exhibit a higher degree of excitement and happiness than positive movie reviews.

\subsubsection{Improving Sentiment Prediction}

The mood manifold captures most of the information for predicting movie review scores or restaurant review scores. Some useful information for review prediction, however, is not captured within the mood manifold. This applies in particular to phrases that are relevant to the review scores, and yet convey no emotional contents. Examples include (in the case of movie reviews) \texttt{Oscar}, \texttt{Shakespearean}, and \texttt{\$300M}.

We thus propose to combine the bag of words TF representation with the mood manifold within a linear regression setting. We regularize the model using a group lasso regularization \citep{Yuan2006}, which performs implicit parameter selection by encouraging sparsity
\vspace{-1.5em}
\begin{align*}
  \argmin_w \frac{1}{n} \sum_{i=1}^n (w_1^T x^{(i)} + w_2^T z^{(i)} - y^{(i)})^2 \\ + \lambda(||w_1||_2 + \lambda_2||w_2||_2).
\vspace{-1em}
\end{align*}
Above, $z^{(i)}$ is the projection of $x^{(i)}$ on the mood manifold, and  $\lambda$ and $\lambda_2$ are regularization parameters. The regularization parameters was determined on performance on validation set and fixed throughout all experiments.

Figure~\ref{fig:review_prediction} shows the test $L_2$ prediction error of our method and baseline (ridge regression trained on the original TF features) as a function of the train set size. The group lasso regression performs consistently better than regression on the original features. The advantage obtained from the mood manifold representation decays with the train set size, which is consistent with statistical theory. In other words, when the train set is relatively small, the mood manifold improves sentiment prediction substantially.

We also compared sentiment prediction using the bag of words features and sentiment prediction using the mood manifold exclusively. The mood manifold regression performs better than bag of words regression for smaller train set sizes but worse for larger train set sizes.

\section{Summary and Discussion}

In this paper, we introduced a continuous representation for human emotions $Z$ and constructed a statistical model connecting it to documents $X$ and to a discrete set of emotions $Y$. Our fitted model bears close similarities to models developed in the psychological literature, based on human survey data. 
The approach of this paper may also be generalized sequentially e.g., \citep{Mao2007b,Mao2009a} and geometrically e.g., \citep{Lebanon2003b,Lebanon2009a,Lebanon2004b,Lebanon2005a,Lebanon2005c,Dillon2007}.

Several attempts were recently made at inferring insights from social media or news data through sentiment prediction. Examples include tracking public opinion \citep{Oconner2010}, estimating political sentiment \citep{Taddy2010}, and correlating sentiment with the stock market \citep{Gilbert2010}. It is likely that the current multivariate view of emotions will help make progress on these important and challenging tasks.

\section*{Acknowledgments}

The authors thank Joonseok Lee, Jeffrey Cohn, and Bo Pang for their comments on the paper. The research was funded in part by NSF grant IIS-0906550.

\clearpage

\bibliographystyle{plainnat}
\bibliography{../../share/externalPapers,../../share/groupPapers}

\begin{thebibliography}{40}
\providecommand{\natexlab}[1]{#1}
\providecommand{\url}[1]{\texttt{#1}}
\expandafter\ifx\csname urlstyle\endcsname\relax
  \providecommand{\doi}[1]{doi: #1}\else
  \providecommand{\doi}{doi: \begingroup \urlstyle{rm}\Url}\fi

\bibitem[Alm(2008)]{Alm2008}
E.C.O. Alm.
\newblock \emph{Affect in text and speech}.
\newblock ProQuest, 2008.

\bibitem[Das and Chen(2007)]{Das2007}
S.~Das and M.~Chen.
\newblock Yahoo! for amazon: Sentiment extraction from small talk on the web.
\newblock \emph{Management Science}, 53\penalty0 (9):\penalty0 1375--1388,
  2007.

\bibitem[Dillon et~al.(2007)Dillon, Mao, Lebanon, and Zhang]{Dillon2007}
J.~Dillon, Y.~Mao, G.~Lebanon, and J.~Zhang.
\newblock Statistical translation, heat kernels, and expected distances.
\newblock In \emph{Uncertainty in Artificial Intelligence}, pages 93--100.
  {AUAI} Press, 2007.

\bibitem[Ganu et~al.(2009)Ganu, Elhadad, and Marian]{Ganu2009}
G.~Ganu, N.~Elhadad, and A.~Marian.
\newblock Beyond the stars: Improving rating predictions using review text
  content.
\newblock In \emph{12th International Workshop on the Web and Databases}.
  Citeseer, 2009.

\bibitem[G{\'e}n{\'e}reux and Evans(2006)]{Genereux2006}
M.~G{\'e}n{\'e}reux and R.~Evans.
\newblock Distinguishing affective states in weblog posts.
\newblock In \emph{AAAI Spring Symposium on Computational Approaches to
  Analyzing Weblogs}, pages 40--42, 2006.

\bibitem[Gilbert and Karahalios(2010)]{Gilbert2010}
E.~Gilbert and K.~Karahalios.
\newblock Widespread worry and the stock market.
\newblock In \emph{Proceedings of the International Conference on Weblogs and
  Social Media}, pages 229--247, 2010.

\bibitem[Holzman and Pottenger(2003)]{Holzman2003}
L.~Holzman and W.~Pottenger.
\newblock Classification of emotions in internet chat: An application of
  machine learning using speech phonemes.
\newblock Technical report, Technical Report LU-CSE-03-002, Lehigh University,
  2003.

\bibitem[Kennedy and Inkpen(2006)]{Kennedy2006}
A.~Kennedy and D.~Inkpen.
\newblock Sentiment classification of movie reviews using contextual valence
  shifters.
\newblock \emph{Computational Intelligence}, 22\penalty0 (2, Special Issue on
  Sentiment Analysis)):\penalty0 110--125, 2006.

\bibitem[Keshtkar and Inkpen(2009)]{Keshtkar2009}
F.~Keshtkar and D.~Inkpen.
\newblock Using sentiment orientation features for mood classification in
  blogs.
\newblock In \emph{IEEE International Conference on Natural Language Processing
  and Knowledge Engineering, 2009. NLP-KE 2009}, pages 1--6, 2009.

\bibitem[Lafferty and Lebanon(2005)]{Lafferty2005a}
J.~Lafferty and G.~Lebanon.
\newblock Diffusion kernels on statistical manifolds.
\newblock \emph{Journal of Machine Learning Research}, 6:\penalty0 129--163,
  2005.

\bibitem[Larsen and Diener(1992)]{Larsen1992}
R.~J. Larsen and E.~Diener.
\newblock Promises and problems with the circumplex model of emotion.
\newblock \emph{Review of Personality and Social Psychology}, 13\penalty0
  (13):\penalty0 25--59, 1992.

\bibitem[Lebanon(2003)]{Lebanon2003b}
G.~Lebanon.
\newblock Learning {Riemannian} metrics.
\newblock In \emph{Proc. of the 19th Conference on Uncertainty in Artificial
  Intelligence}. {AUAI Press}, 2003.

\bibitem[Lebanon(2005{\natexlab{a}})]{Lebanon2005a}
G.~Lebanon.
\newblock Axiomatic geometry of conditional models.
\newblock \emph{IEEE Transactions on Information Theory}, 51\penalty0
  (4):\penalty0 1283--1294, 2005{\natexlab{a}}.

\bibitem[Lebanon(2005{\natexlab{b}})]{Lebanon2005b}
G.~Lebanon.
\newblock \emph{Riemannian Geometry and Statistical Machine Learning}.
\newblock PhD thesis, Carnegie Mellon University, Technical Report
  CMU-LTI-05-189, 2005{\natexlab{b}}.

\bibitem[Lebanon(2005{\natexlab{c}})]{Lebanon2005c}
G.~Lebanon.
\newblock Information geometry, the embedding principle, and document
  classification.
\newblock In \emph{Proc. of the 2nd International Symposium on Information
  Geometry and its Applications}, pages 101--108, 2005{\natexlab{c}}.

\bibitem[Lebanon(2009)]{Lebanon2009a}
G.~Lebanon.
\newblock Axiomatic geomtries for text documents.
\newblock In P.~Giblisco, E.~Riccomagno, M.~P. Rogantin, and H.~P. Wynn,
  editors, \emph{Algebraic and Geometric Methods in Statistics}. Cambridge
  University Press, 2009.

\bibitem[Lebanon and Lafferty(2004)]{Lebanon2004b}
G.~Lebanon and J.~Lafferty.
\newblock Hyperplane margin classifiers on the multinomial manifold.
\newblock In \emph{Proc. of the 21st International Conference on Machine
  Learning}. Morgan Kaufmann Publishers, 2004.

\bibitem[Mao and Lebanon(2007)]{Mao2007b}
Y.~Mao and G.~Lebanon.
\newblock Isotonic conditional random fields and local sentiment flow.
\newblock In \emph{Advances in Neural Information Processing Systems 19}, pages
  961--968, 2007.

\bibitem[Mao and Lebanon(2009)]{Mao2009a}
Y.~Mao and G.~Lebanon.
\newblock Generalized isotonic conditional random fields.
\newblock \emph{Machine Learning}, 77\penalty0 (2-3):\penalty0 225--248, 2009.

\bibitem[Mishne(2005)]{Mishne2005}
G.~Mishne.
\newblock Experiments with mood classification in blog posts.
\newblock In \emph{1st Workshop on Stylistic Analysis Of Text For Information
  Access}, 2005.

\bibitem[Mohammad and Turney(2011)]{Mohammad2011}
S.~M. Mohammad and P.~D. Turney.
\newblock Crowdsourcing a word--emotion association lexicon.
\newblock \emph{Computational Intelligence}, 59\penalty0 (000):\penalty0 1--24,
  2011.

\bibitem[Na et~al.(2004)Na, Sui, Khoo, Chan, and Zhou]{Na2004}
J.~Na, H.~Sui, C.~Khoo, S.~Chan, and Y.~Zhou.
\newblock Effectiveness of simple linguistic processing in automatic sentiment
  classification of product reviews.
\newblock \emph{Advances in Knowledge Organization}, 9:\penalty0 49--54, 2004.

\bibitem[Nakagawa et~al.(2010)Nakagawa, Inui, and Kurohashi]{Nakagawa2010}
T.~Nakagawa, K.~Inui, and S.~Kurohashi.
\newblock Dependency tree-based sentiment classification using crfs with hidden
  variables.
\newblock In \emph{Human Language Technologies: The 2010 Annual Conference of
  the North American Chapter of the Association for Computational Linguistics},
  pages 786--794. Association for Computational Linguistics, 2010.

\bibitem[O'Connor et~al.(2010)O'Connor, Balasubramanyan, Routledge, and
  Smith]{Oconner2010}
B.~O'Connor, R.~Balasubramanyan, B.~Routledge, and N.~Smith.
\newblock From tweets to polls: Linking text sentiment to public opinion time
  series.
\newblock In \emph{Proceedings of the International AAAI Conference on Weblogs
  and Social Media}, pages 122--129, 2010.

\bibitem[Pang and Lee(2005)]{Pang2005}
B.~Pang and L.~Lee.
\newblock Seeing stars: Exploiting class relationships for sentiment
  categorization with respect to rating scales.
\newblock In \emph{Proceedings of the 43rd Annual Meeting on Association for
  Computational Linguistics}, 2005.

\bibitem[Pang and Lee(2008)]{Pang2008}
B.~Pang and L.~Lee.
\newblock Opinion mining and sentiment analysis.
\newblock \emph{Found. Trends Inf. Retr.}, 2:\penalty0 1--135, 2008.
\newblock ISSN 1554-0669.

\bibitem[Quan and Ren(2009)]{Quan2009}
C.~Quan and F.~Ren.
\newblock Construction of a blog emotion corpus for chinese emotional
  expression analysis.
\newblock In \emph{Proceedings of the 2009 Conference on Empirical Methods in
  Natural Language Processing: Volume 3-Volume 3}, pages 1446--1454.
  Association for Computational Linguistics, 2009.

\bibitem[Rubin et~al.(2004)Rubin, Stanton, and Liddy]{Rubin2004}
V.~Rubin, J.~Stanton, and E.~Liddy.
\newblock Discerning emotions in texts.
\newblock In \emph{The AAAI Symposium on Exploring Attitude and Affect in Text
  (AAAI-EAAT)}, 2004.

\bibitem[Russell(1979)]{Russell1979}
J.~A. Russell.
\newblock Affective space is bipolar.
\newblock \emph{Journal of personality and social psychology}, 37\penalty0
  (3):\penalty0 345, 1979.

\bibitem[Russell(1980)]{Russell1980}
J.~A. Russell.
\newblock A circumplex model of affect.
\newblock \emph{Journal of personality and social psychology}, 39\penalty0
  (6):\penalty0 1161, 1980.

\bibitem[Shaver et~al.(1987)Shaver, Schwartz, Kirson, and O'connor]{Shaver1987}
P.~Shaver, J.~Schwartz, D.~Kirson, and C.~O'connor.
\newblock Emotion knowledge: Further exploration of a prototype approach.
\newblock \emph{Journal of personality and social psychology}, 52\penalty0
  (6):\penalty0 1061, 1987.

\bibitem[Socher et~al.(2011)Socher, Pennington, Huang, Ng, and
  Manning]{Socher2011}
R.~Socher, J.~Pennington, E.~H. Huang, A.~Y. Ng, and C.~Manning.
\newblock Semi-supervised recursive autoencoders for predicting sentiment
  distributions.
\newblock In \emph{Proceedings of the Conference on Empirical Methods in
  Natural Language Processing}, pages 151--161. Association for Computational
  Linguistics, 2011.

\bibitem[Strapparava and Mihalcea(2008)]{Strapparava2008}
C.~Strapparava and R.~Mihalcea.
\newblock Learning to identify emotions in text.
\newblock In \emph{Proceedings of the 2008 ACM symposium on Applied computing},
  pages 1556--1560. ACM, 2008.

\bibitem[Strapparava and Valitutti(2004)]{Strapparava2004}
C.~Strapparava and A.~Valitutti.
\newblock Wordnet-affect: an affective extension of wordnet.
\newblock In \emph{Proceedings of LREC}, volume~4, pages 1083--1086. Citeseer,
  2004.

\bibitem[Taddy(2010)]{Taddy2010}
M.~Taddy.
\newblock {Inverse Regression for Analysis of Sentiment in Text}.
\newblock \emph{Arxiv preprint \texttt{arXiv:1012.2098}}, 2010.

\bibitem[Tellegen et~al.(1999)Tellegen, Watson, and Clark]{Tellegen1999}
A.~Tellegen, D.~Watson, and L.~A. Clark.
\newblock {On The Dimensional and Hierarchical Structure of Affect}.
\newblock \emph{Psychological Science}, 10\penalty0 (4):\penalty0 297--303,
  1999.
\newblock ISSN 1467-9280.

\bibitem[Watson and Tellegen(1985)]{Watson1985}
D.~Watson and A.~Tellegen.
\newblock Toward a consensual structure of mood.
\newblock \emph{Psychological bulletin}, 98\penalty0 (2):\penalty0 219--235,
  September 1985.
\newblock ISSN 0033-2909.

\bibitem[Watson et~al.(1988)Watson, Clark, and Tellegen]{Watson1988}
D.~Watson, L.~A. Clark, and A.~Tellegen.
\newblock Development and validation of brief measures of positive and negative
  affect: the panas scales.
\newblock \emph{Journal of personality and social psychology}, 54\penalty0
  (6):\penalty0 1063, 1988.

\bibitem[Yik et~al.(2011)Yik, Russell, and Steiger]{Yik2011}
M.~Yik, J.~Russell, and J.~Steiger.
\newblock A 12-point circumplex structure of core affect.
\newblock \emph{Emotion}, 11\penalty0 (4):\penalty0 705, 2011.

\bibitem[Yuan and Lin(2006)]{Yuan2006}
M.~Yuan and Y.~Lin.
\newblock Model selection and estimation in regression with grouped variables.
\newblock \emph{Journal of the Royal Statistical Society: Series B (Statistical
  Methodology)}, 68, 2006.

\end{thebibliography}

\end{document}